# Structural Health Monitoring Using Neural Network Based Vibrational System Identification

Donald A. Sofge, Vice-President
NeuroDyne Inc.
One Kendall Square
Cambridge, Massachusetts 02139, USA
e-mail: sofge@ai.mit.edu

## Abstract

Composite fabrication technologies now provide the means for producing high-strength, low-weight panels, plates, spars and other structural components which use embedded fiber optic sensors and piezoelectric transducers. These materials, often referred to as *smart structures*, make it possible to sense internal characteristics, such as delaminations or structural degradation. In this effort we use neural network based techniques for modeling and analyzing dynamic structural information for recognizing structural defects. This yields an adaptable system which gives a measure of structural integrity for composite structures.[*]

## Background

Composite fabrication technologies now provide a means for producing high-strength, low weight structural components with structural and material properties which can be tailored to use in specific target environments (e.g. stiffness con-straints, durability, temperature and corrosion resistance). However, these composite structures may degrade due to improper manufacture (improper cure cycle, fiber misalignment, foreign object debris, delaminations, etc.), duty cycle wear, impacts, and material corrosion (e.g. due to moisture, fuel or chemical absorption). This structural degradation may not be detectable through visual inspection.

Fiber optic sensing technology provides a means for sensing various structural properties such as stress, strain, and elasticity when these sensors are mounted on, or embedded within, a material structure. Current composite fabrication methods have been demonstrated which allow fiber optic sensors to be embedded within a composite structure during manufacture of that structure. These sensors maintain their viability throughout the cure cycle and various other production stages, and thereafter may *potentially* be used for structural health monitoring. Although a large variety of embedded fiber optic sensors have been explored, we focus on perhaps the simplest and most reliable method: the fiber optic strain sensor.

Another class of devices which may be embedded in, or mounted on, composite structures is piezoelectric transducers, or PZTs. A composite which uses a number of embedded PZTs may utilize them for vibration sensing or generation, but any given transducer may be used only for one of these purposes at a given time. Other than for exciting the vibrational modes of a structure, the use of PZTs for vibrational control is beyond the scope of this paper.

The key to understanding how embedded fiber optic sensors (in particular, strain sensors) may be used to determine structural integrity is the understanding of how various types of degradation correlate with changes in the structural dynamics of the localized area of the composite structure. Impacts, material defects, and wear all affect the structural dynamics of the composite (hence the need for structural integrity monitoring) in a fashion which is measurable based upon the response of the structure to normal everyday use. Neural networks may be trained to recognize the symptoms of structural degradation based upon changes in the dynamic response of the composite part, and to correlate these symptoms with their root causes (e.g. impact damage, delaminations, duty cycle wear).

Therefore, the task for determining structural integrity using embedded fiber optic sensors may be cast as (1) determine the structural dynamics of the localized area, (2) correlate dynamic information to local structural degradation, and (3) integrate this knowledge (in real time) over the entire vehicle to provide input to a vehicle health monitoring system. The first task has long been studied from the viewpoint of systems theory, controls, and system identification theory.

Existing tactical combat aircraft (e.g. F-15E Eagle) have rudimentary overload warning systems which provide an audible tone to the pilot when he exceeds the design limit load of the structure. What is required for health monitoring in smart structures for high performance aircraft is a much more general, sophisticated, and automated system which provides quantitative analysis of the overall structure's state. Additionally, periodic non-destructive inspection (NDI) inspections of the structure (ultrasonic imaging, thermography, eddy current, etc) are time/labor consuming, result in decreased vehicle survivability, and may not even provide coverage of inaccessible areas. In terms of inspection, smart structure technology would greatly reduce NDI inspection requirements by re-ducing the frequency of NDI and by detecting new problem areas autonomously.

---
[*] This material is based upon work supported by the National Science Foundation under award number III-9360372, and was supported in part by NASA under award NASW-4921.

## System Identification Methodology

Neural networks present a powerful tool for the identification of multivariable nonlinear systems. The full analytical model is often either unavailable, or available but unmanageable. The situation of a not fully observable system may also arise when we do not fully understand a system. Consequently, we may not know that certain parameters need to be measured even if they are available for measurement. The identification task can be further compounded if the nonlinear system is a large scale system where the common supposition of centrality in system analysis is no longer valid (e.g. identification of large flexible structures). In our approach, predictive neural networks have been used within a conventional modal analysis framework and shown to yield solid results for time-varying multi-mode system identification of a composite structure in the form of a cantilevered beam.

Perhaps the most promising capability of neural network algorithms is the ability to learn, both off-line and on-line. A common oversight in identifying the structural dynamics for composites using bonded or embedded sensors is the assumption of a perfect composite structure. There often exist significant discrepancies between mathematical models and the actual physical structure due to structural flaws such as delaminations. These flaws can change the natural modal response and alter the sensor response. Even if a structure can be manufactured perfectly, damage to the structure while in use may corrupt sensor measurements or alter the modal behavior. The solution is an adaptive or learning system that can model the dynamics of structures with damage induced during manufacture or after implementation.

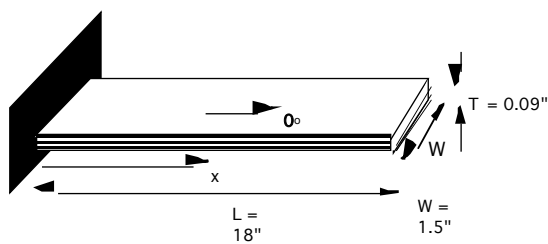

**Figure 1.** Laminated beam (18 plies). The finite element model has 37 nodes, 36 elements.

## Predictive Neural Network for Forward Modeling of Structural Dynamics

Because of the time varying characteristic of the displacement of a vibrating composite structure, the neural network model should have time series prediction capability. If our goal is to predict x(k), we will need input of x(k-1), x(k-2), --- x(k-n) where k indicates time stage k, and n is the window size.

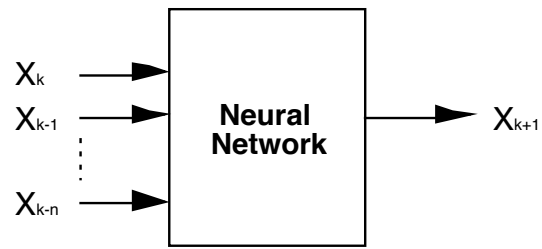

**Figure 2.** Predictive Network Model Using Tapped Delay Line

The system may be identified using random excitations along the composite structure. This in effect models the Transient Dynamic Response (TDR) of the vehicle. The TDR results from the forces and loads impinged upon the vehicle (or structure) during normal operation. This will make it possible to perform system identification passively, rather than requiring embedded actuators or larger vibration devices such as shakers to excite the modes of the structure.

For a cantilevered beam, the first 4 modes generally yield reasonable information for vibration analysis. The frequencies of the first 4 modes can be approximated based on our knowledge of the beam material. The first 4 modes in the simulation code are shown below:

| mode #(i) | $\omega_i$ | $f_i = \omega_i/2\pi$ |
|---|---|---|
| 1 | 10 | 1.6 |
| 2 | 62 | 9.9 |
| 3 | 175 | 27.7 |
| 4 | 343 | 54.4 |

The frequency information can be used to give us a range of values for the size of the neural network. The highest frequency of interest is 54.4 hertz. The window should cover 1/2 of the wave length of $\omega_n$ to give us enough information for autoregression. If the window size (number of samples) is too large, an averaging effect starts to set in, and the network will lose sensitivity to detail. At a sampling rate of 1000 hz, a target window size n should be about 9 which is 1000 divided by (2 x 54). At a higher sampling frequency, we were able to use a larger n.

A second factor affecting the number of samples in the window is the order of the system. A higher order system will need a larger number of window size for input. The order of the system also dictates the number of hidden nodes. For a given application article (such as a composite beam), a series of tests may be conducted to select the optimal network architecture.

A series of tests was run using the composite beam simulation code to generate data, and then modeling the beam dynamics using a 2-hidden-layer sigmoidal network.. This experience has suggested that the optimal 2-hidden-layer network uses a window size of 9 and 25 hidden nodes per layer, however this may not be true for a damaged panel or beam. The output layer uses linear nodes. Further work under development (not yet ready for publication at the time of this writing) involves using a linear network to learn the beam dynamics, and then using a second sigmoidal network to analyze the weights of this network (which are the coefficients of a linear system of equations which describe the vibrational dynamics) to diagnose structural defects.

Modeling the effects of structural degradation on the beam required modifications to the beam response simulation software to simulate the effects of damage. Finite Element Analysis (FEM) was used to model 3 different levels of damage in the form of delaminations, which in the FEM code resulted in stiffness knockdowns at specific locations in the finite element model. In addition, two different lengths of delaminations were simulated (1" and 2"), and each of these scenarios was tested in each of 10 different locations along the beam. This resulted in 60 different damage configurations, plus one for which no damage was present. For each of the 61 beam configurations the first four vibrational modes were identified. This information was then incorporated into the composite beam simulation code, with an interface for the user to select desired level and location of damage only the beam. The beam simulator then was used to simulate the dynamics of damaged composite beams.

The predictive neural network was then used to identify the new beam dynamics. Once a basic model of the structural dynamics has been achieved, it may be gradually updated over time, or as damage occurs, based upon new sensor response data and the transient dynamic response (TDR) of the structure. We have found it extremely useful to use a very low learning rate for such on-line learning tasks.

### *Structural Integrity based upon Transient Dynamic Response*

Modal analysis performed on the forward model (neural network) of the beam dynamics provides a basis upon which to predict local structural integrity. This information may be augmented with real-time feedback from an embedded fiber optic sensor or PZT which measure the response of the structure to a random transient excitation. We refer to this as the Transient Dynamic Response (TDR) of the structure. Hence, a neural network based classifier may integrate both system identification knowledge as well as real-time embedded sensor feedback to make a real-time determination of local structural integrity. This network may then be used as a basic "smart node" from which information is then fed into a higher level knowledge integrator which provides sensor fusion, redundancy, and robustness against sensor damage.

In a previous related effort researchers from MDA and NeuroDyne demonstrated that statically trained neural networks can be used to classify impact levels on composite panels based upon actual impact test data and embedded sensor signals (Figure 3).

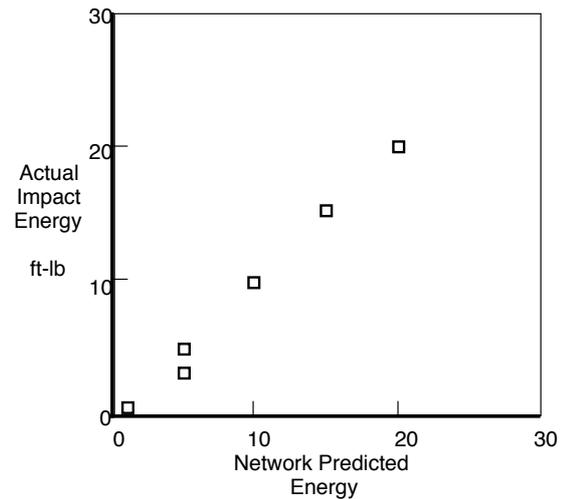

**Figure 3.** Neural Network based Impact Prediction

Impacts, material defects, and wear all affect the structural dynamics of the composite in a fashion which is measurable based upon the response of the structure to normal everyday use. Neural networks may be trained to recognize the symptoms of structural degradation based upon changes in the dynamic response of the composite part, and to correlate these symptoms with their root causes (e.g. impact damage, delaminations, duty cycle wear). The neural networks used in this effort are multi-layer perceptrons (MLPs), although we are finding new ways to integrate these with linear networks for vibrational modeling, with the MLP networks used for damage classification. The classification network may then classify the degree and type of the damage present.

In future efforts these networks will be trained off-line on data gathered from a composite panel with embedded fiber optic sensors. An objective of this study is to identify measures of structural integrity appropriate for application of composite structure technology to vehicles such as satellites or deep space probes, and to define a neural network architecture for classifying these defects. In further work we will fabricate composite beams with embedded fiber sensors, and with carefully controlled defects in the form of delaminations. The effects of these various defects will be verified, and the classifier architecture will be adjusted suitably to reflect these damage conditions.

**Effect of Delamination on Natural Frequencies**

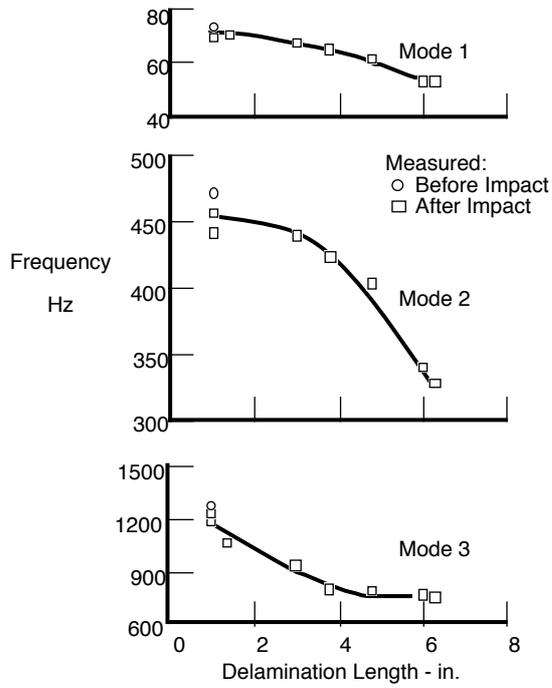

**Figure 4.** Damage Assessment via Dynamic Response Change

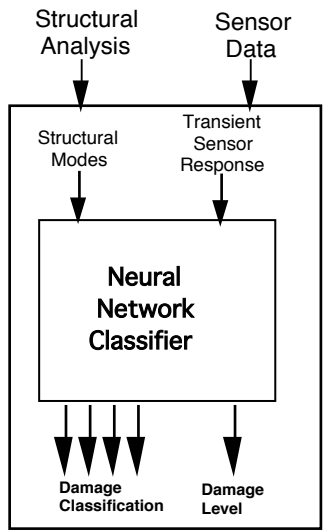

**Figure 5.** Damage Classifier